\documentclass{article}
\usepackage{graphicx}
\usepackage{subcaption}
\usepackage{multirow}
\usepackage{array}

    \usepackage[preprint]{neurips_2025}

\usepackage[utf8]{inputenc} 
\usepackage[T1]{fontenc}    
\usepackage{hyperref}       
\usepackage{url}            
\usepackage{booktabs}       
\usepackage{amsfonts}       
\usepackage{nicefrac}       
\usepackage{microtype}      
\usepackage{xcolor}         
\usepackage{amsmath} 
\usepackage{multirow} 
\usepackage{adjustbox}
\usepackage{wrapfig}

\title{FullDiT2: Efficient In-Context Conditioning \\for Video Diffusion Transformers }

\author{%
  Xuanhua He$^{1}$\footnotemark[1], Quande Liu$^{2}$\footnotemark[2], Zixuan Ye$^{1}$, Weicai Ye$^{2}$, Qiulin Wang$^{2}$, 
  \\ \textbf{Xintao Wang}$^{2}$, \textbf{Qifeng Chen}$^{1}$, \textbf{Pengfei Wan}$^{2}$, \textbf{Di Zhang}$^{2}$, \textbf{Kun Gai}$^{2}$\\
  \textsuperscript{1}The Hong Kong University of Science and Technology \\
  \textsuperscript{2}Kuaishou Technology
}

\begin{document}
\maketitle
\let\thefootnote\relax\footnotetext{
$^*$ Work done during an internship at KwaiVGI, Kuaishou Technology. $^\dagger$ Corresponding authors
}

\begin{abstract}

Fine-grained and efficient controllability on video diffusion transformers has raised increasing desires for the applicability. Recently, In-context Conditioning emerged as a powerful paradigm for unified conditional video generation, which enables diverse controls by concatenating varying context conditioning signals with noisy video latents into a long unified token sequence and jointly processing them via full-attention, e.g., FullDiT. Despite their effectiveness, these methods face quadratic computation overhead as task complexity increases, hindering practical deployment. In this paper, we study the efficiency bottleneck neglected in original in-context conditioning video generation framework. We begin with systematic analysis to identify two key sources of the computation inefficiencies: the inherent redundancy within context condition tokens and the computational redundancy in context-latent interactions throughout the diffusion process. Based on these insights, we propose FullDiT2, an efficient in-context conditioning framework for general controllability in both video generation and editing tasks, which innovates from two key perspectives. Firstly, to address the token redundancy in context conditions, FullDiT2 leverages a dynamical token selection mechanism to adaptively identity important context tokens, reducing the sequence length for unified full-attention. Additionally, a selective context caching mechanism is devised to minimize redundant interactions between condition tokens and video latents throughout the diffusion process. Extensive experiments on six diverse conditional video editing and generation tasks demonstrate that FullDiT2 achieves significant computation reduction and 2-3 times speedup in averaged time cost per diffusion step, with minimal degradation or even higher performance in video generation quality. The project page is at \href{https://fulldit2.github.io/}{https://fulldit2.github.io/}.

\end{abstract}

\section{Introduction}
Diffusion transformers (DiTs) have achieved remarkable progress in video generation~\cite{yang2024cogvideox,wang2025wan,peng2025open}. To provide flexible control over the generated videos, effective controlling mechanisms have garnered increasing attention. Traditionally, methods have relied on adapter-based techniques~\cite{he2024id,ye2024stylemaster,liu2023stylecrafter}, such as ControlNet~\cite{che2024gamegen,guo2024sparsectrl} or IP-Adapter~\cite{ma2024followyouremoji,liu2023stylecrafter}, to inject various condition signals into diffusion models. While these methods have shown promising progress, they are typically task-specific, necessitating independent modules for each task and leading to noticeable module conflicts in multi-condition controls~\cite{peng2024controlnext}.

In response to these limitations, a new conditioning mechanism has emerged, aiming to unify various controllable video generation tasks by leveraging the native representation learning capabilities of DiTs~\cite{tan2024ominicontrol,lin2025omnihuman,guo2025long,guo2025long}, representatively as FullDiT~\cite{ju2025fulldit}. The core idea is to tokenize multi-modal control signals and concatenate them with the noisy video latents into a single cohesive token sequence, jointly modeled using full-attention. 
Unlike previous approaches requiring auxiliary adapters, these methods simply reuse the original attentions of DiTs to model the complex interaction between context condition tokens and video latents, with no architecture modification. For clarity, we refer to this mechanism as "in-context conditioning (ICC)", considering its nature to perform varying controllable video generation tasks by simply referring the diverse condition tokens "in context". Such design shows superior flexibility in various conditioning video generation scenarios.  

However, despite its benefits, this unified approach face critical challenges: the computation cost as the task complexity escalates. For example, in video generation, the condition signals are generally long-sequential, such as pose or camera sequences~\cite{he2024cameractrl,ma2024followyourpose}, which substantially increase the token numbers within the unified sequence length. When extended to video editing tasks, the sequence length becomes even longer, as multiple conditions would be jointly considered (e.g., the reference source video and other condition signals)~\cite{bai2025recammaster}. Consequently, as the task complexity and token length increases, such in-context conditioning mechanism would faces exponential-growth computation overhead, hindering its extendability in complex video conditioning and editing tasks.

\begin{figure}[t]
    \centering
    \includegraphics[width=0.9\linewidth]{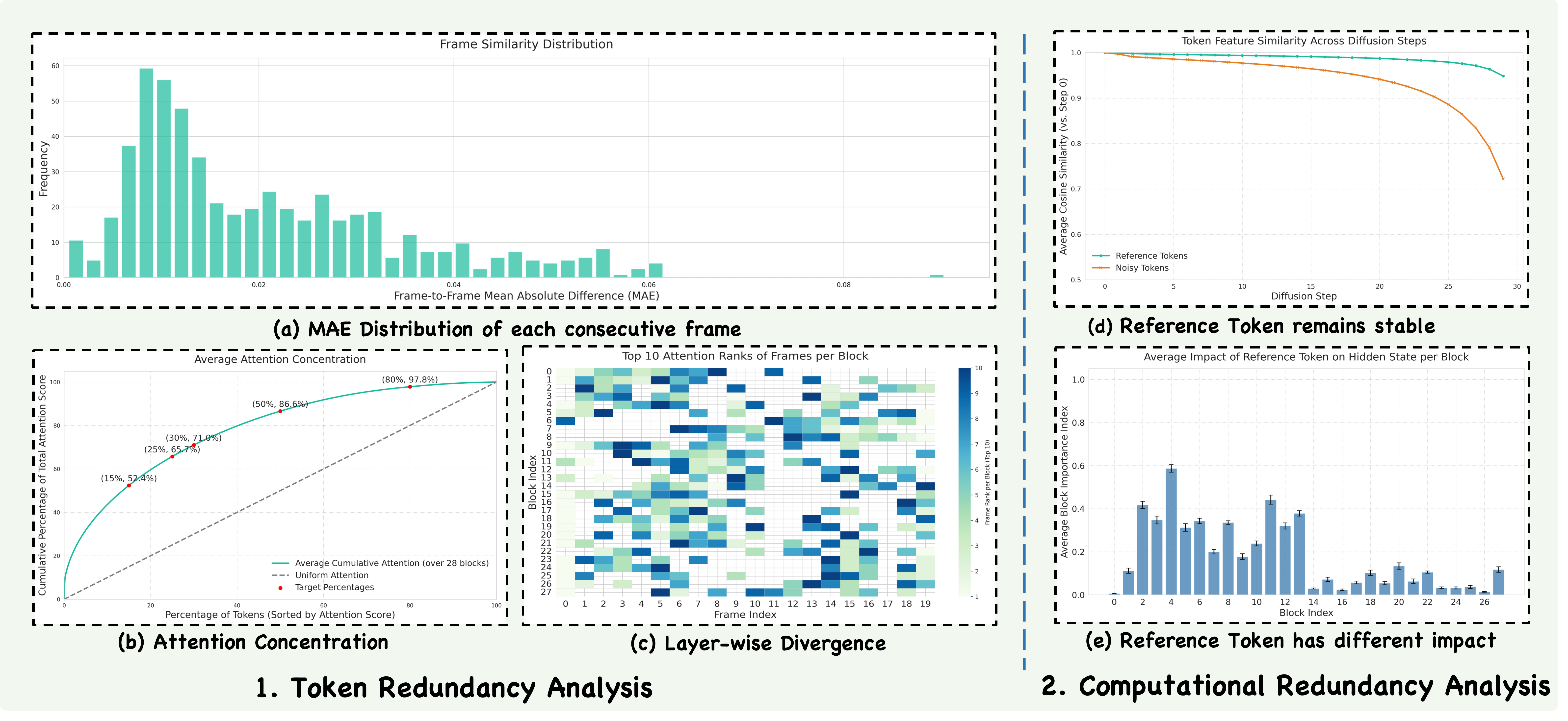}
    \caption{Empirical Analysis of Redundancy in In-Context Conditioning. \textbf{Token Redundancy} in video object editing. (a) Absolute difference across the averaged tokens of each consecutive frame. (b) The Average Attention Concentration curve shows that ~50\% of reference tokens capture >85\% of cumulative attention score from noisy latents (averaged over blocks). (c) Layer-wise Divergence heatmap indicates that different blocks focus attention on different reference tokens (grouped by frames). \textbf{Computational Redundancy} in video recamera. (d) Average cosine similarity shows reference token features remain stable across diffusion steps compared to noisy tokens. (e) The Block Importance Index reveals varied impact of reference tokens across different network layers.}
    \label{fig:video_redudant}
\vspace{-5mm}
\end{figure}

In this paper, we therefore aim to address a challenging problem, how to effectively address the computation burden from ICC under the complex conditioning scenarios for video generation and editing. To achieve efficient condition control, we start from analytical experiments to identify the computation redundancy of FullDiT baseline from two orthogonal yet critical perspectives, and draw the following observations (detailed analysis in Sec~\ref{sec:ana}): 

\quad\quad\textbf{Token redundancy of context condition signals:} In ICC-driven video generation and editing tasks, we observe a significant redundancy within the conditional tokens, as evidenced by the long-tail patterns in the differences across averaged tokens of consecutive frames in Figure~\ref{fig:video_redudant} (a). This redundancy results in inefficient computation during full self-attention operations. More importantly, we observe that the network blocks dynamically focus on different subsets of context tokens throughout the diffusion process (Figure~\ref{fig:video_redudant} (c)), further highlighting the redundancy of previous ICC operations and the variability in its focus of attention.

\quad\quad\textbf{Computation Redundancy between Context-Latent Interaction:} The interaction between context tokens and noise latent presents high redundancies due to their repetitive bi-directional full-attention, across both step-level and layer-level. Despite the noisy latents are dynamically updated, we observe in the experiment of Figure~\ref{fig:video_redudant} (d) that the context tokens remain relatively static throughout the diffusion steps, revealing the severe step-wise computational redundancy in the continuous update of context tokens. Step further in Figure~\ref{fig:video_redudant} (e), our layer-wise analysis uncovers an unbalanced influence of context tokens on video latent across different transformer blocks, highlighting the inefficiency of previous full-block interaction between context tokens and video latent.

\begin{figure}
    \centering
    \includegraphics[width=0.9\linewidth]{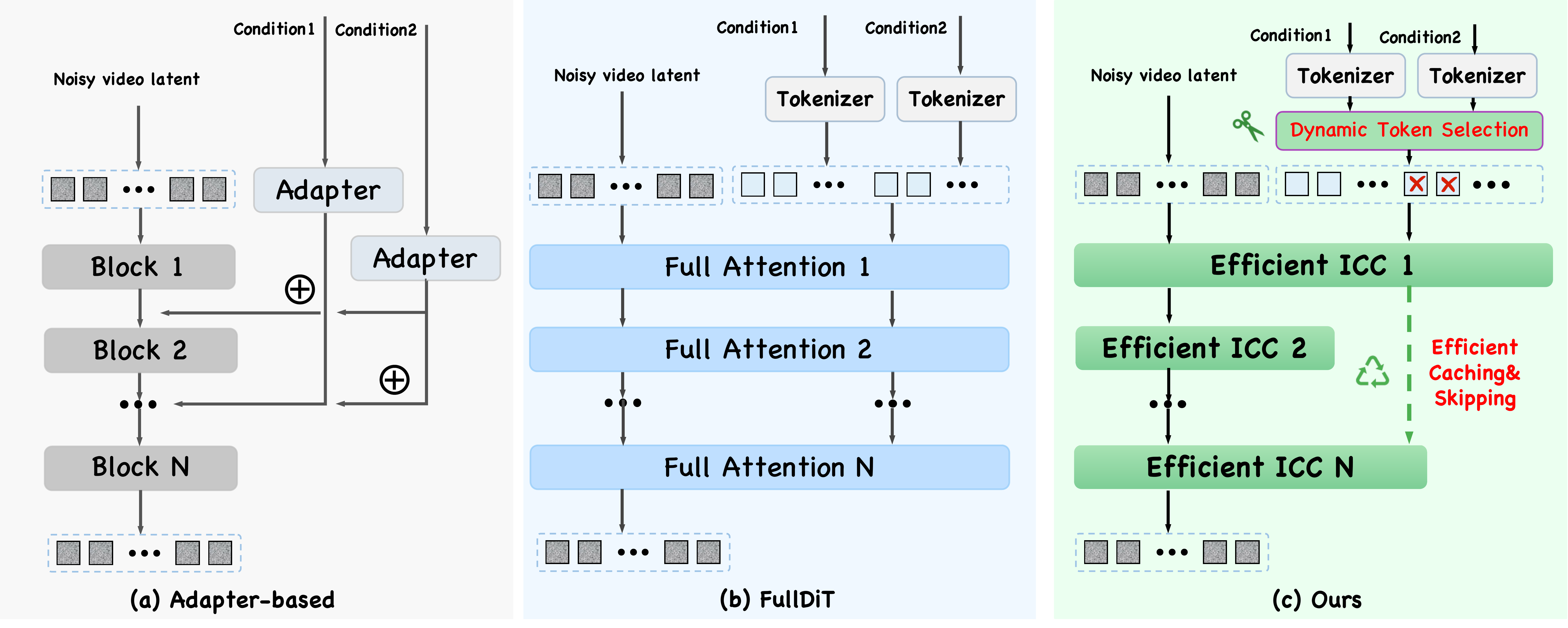}
    \caption{Comparison of our FullDiT2 with adapter-based methods and Full-DiT.}
    \label{fig:cmap1}
\vspace{-8mm}
\end{figure}
Based on these insights above, we finally propose FullDiT2, and efficient in context-condition framework for unified conditioning of various video generation and editing tasks. As shown in Figure~\ref{fig:cmap1}, FullDiT2 inherits the context conditioning mechanism to jointly model the long token sequence composed of multimodal condition tokens and video latent, yet showing two key innovations that makes it simple yet effective: Firstly, to address the token redundancy of context conditions, which lead to exponential computation increase, we propose to dynamically identify a subset of important context tokens, reducing the sequence length for full-attention operations. Given the observations that the context features are stable across steps and their impact on video latents are usually fluctuated, we further devise a selective context caching mechanism, which minimize the redundant computations between the interaction of condition tokens and video latents by effecitvely caching and skipping the context tokens across diffusion step and blocks.

To summarize, the key contribution of this paper are: 

\begin{itemize}
\item We conduct a comprehensive analysis of computation redundancy within the in-context conditioning framework, identifying two critical aspects: token redundancy of context condition signals and computation redundancy between context-latent interactions. It highlights the significant inefficiencies in the current context conditioning-based methods.

\item We propose FullDiT2, an efficient in-context conditioning framework for unified control in video generation and editing tasks.  It effectively enhances the model efficiency by adaptively selecting context tokens to mitigate exponential computation increases in full-attention, while minimizing redundant computations through selective caching and skipping of context tokens across diffusion steps and blocks.

\item  We conduct extensive experiments on six different video editing and controllable video generation tasks. The results demonstrate that FullDiT2 achieves substantial reductions in FLOPs with a 2-3x speedup in averaged time cost per diffusion step, while maintaining or even outperforming the video generation peformance over FullDiT.
\end{itemize}

\section{Related Work}
\subsection{In-context Conditioning Diffusion Transformers}
Recently, researchers have begun exploring the In-Context Conditioning capabilities of Diffusion Transformers, particularly through sequence concatenation approaches for conditioning~\cite{huang2024context,tan2024ominicontrol,wu2025less,li2025visualcloze,song2025insert,mou2025dreamo}. Early explorations, such as Omnicontrol~\cite{tan2024ominicontrol, tan2025ominicontrol2}, demonstrated the potential of concatenating conditioning tokens with the standard noisy latents into a single extended sequence. This unified sequence is then processed jointly using the Transformer's self-attention mechanism, allowing the model to learn conditioning relationships implicitly. A key advantage of this strategy is its elegance: it enables powerful multi-modality control without requiring modifications to the DiT architecture, often leading to improved generation quality and controllability.
Building on this paradigm, numerous methods employing in-context conditioning via concatenation were proposed in the image generation domain~\cite{wu2025less,li2025visualcloze,song2025insert,mou2025dreamo}. Subsequently, this approach was successfully extended to the more challenging video generation domain, yielding impressive results~\cite{ju2025fulldit,lin2025omnihuman,guo2025long}. Notable examples include: OmniHuman~\cite{lin2025omnihuman}, which conditions human video synthesis on various inputs like poses or appearance references; FullDiT~\cite{ju2025fulldit}, enabling controllable text-to-video generation guided by diverse conditions such as depth video or camera trajectories; and LCT~\cite{guo2025long}, which leverages concatenated video shots for tasks like consistent video generation.
\subsection{Efficient Computation of Diffusion Transformers}
Improving the computational efficiency of Diffusion Transformers is crucial for their practical application. Research efforts  usually follow two directions: reducing the steps of sampling process and accelerating the denoising network computation per step. Significant progress has been made in timestep reduction through advanced samplers (e.g., DDIM~\cite{song2020denoising}, DPM-Solvers~\cite{song2020denoising}), distillation techniques~\cite{salimans2022progressive,meng2023distillation}, and consistency models~\cite{luo2023latent}. Orthogonal to these, several strategies exist to accelerate the network computation per step. Techniques like weight quantization~\cite{li2023q}, pruning~\cite{castells2024ld}, and general token merging~\cite{bolya2023token,wu2024importance} modify the model or its inputs. Another line of work reduces redundant computations for the noisy latent pathway ($z_t$) during inference. Methods such as FORA~\cite{selvaraju2024fora}, $\delta$-DiT~\cite{chen2024delta}, Learning-to-Cache~\cite{ma2024learning} achieve this by reusing intermediate features or residuals related to $z_t$ across timesteps or network blocks, effectively minimizing repeated processing of the noisy state. These approaches primarily target the $z_t$ computations. While some methods accelerate separate conditional branches~\cite{cao2025relactrl,peng2024controlnext}, the unique computational characteristics and redundancies arising from reference condition tokens processed within the unified ICC sequence remain largely underexplored. 
\section{FullDiT2}
\subsection{Preliminary: In-context Conditioning for Video Generation and Editing}
Video Diffusion Transformer (DiT)~\cite{peebles2023scalable} models learn the conditional distribution $p(\mathbf{z} | C)$ by reversing a diffusion process. This process gradually adds noise to clean data $\mathbf{z}_1$ to obtain noise $\mathbf{z}_0 \sim \mathcal{N}(\mathbf{0}, \mathbf{I})$. Formulations like Flow Matching~\cite{lipman2022flow} train a network $\mathbf{u}_{\Theta}(\mathbf{z}_t, t, C)$ to predict the velocity $\mathbf{v}_t$ on the path from $\mathbf{z}_0$ to $\mathbf{z}_1$, often via an MSE loss:
\begin{equation}
\mathcal{L} = \mathbb{E}_{t, \mathbf{z}_0, \mathbf{z}_1, C} \left\| \mathbf{u}_{\Theta}(\mathbf{z}_t, t, C) - \mathbf{v}_t \right\|^2
\end{equation}
Generation involves sampling $\mathbf{z}_0$ and integrating the learned velocity $\mathbf{u}_{\Theta}$ using an ODE solver. The network $\mathbf{u}_{\Theta}$ commonly consists of stacked Transformer blocks $\mathrm{g_l}$, each potentially containing full spatio-temporal self-attention ($\mathrm{F^l_{SA}}$), cross-attention ($\mathrm{F^l_{CA}}$), and MLPs ($\mathrm{F^l_{MLP}}$).

The condition $C$ serves to guide both the training and sampling processes. To incorporate $C$, some approaches employ dedicated control branches or adapter modules. However, these methods often necessitate network modifications and may suffer from limited generalization capabilities to diverse conditions or tasks.
A distinct strategy for incorporating the condition $C$ is \textbf{In-Context Conditioning (ICC)}. Unlike methods requiring dedicated control branch or adapter modules, ICC operates directly within the main DiT architecture. It involves concatenating tokens derived from the conditioning signal $\mathbf{c}$ with the noisy latent tokens $\mathbf{z}_t$ into a single, extended sequence:
\begin{equation}
\mathbf{s} = [\mathbf{z}_t;\mathbf{c}]
\end{equation}
where $[\cdot; \cdot]$ denotes sequence concatenation.

This unified sequence $\mathbf{s}$ is then processed jointly by the standard DiT blocks (e.g., $\mathrm{g_l}$). The self-attention mechanisms within these blocks, particularly the full spatio-temporal attention $\mathrm{F^l_{SA}}$ which models both intra-frame and cross-frame dependencies, are responsible for dynamically modeling the complex relationships between the conditioning tokens $\mathbf{c}$ and the noisy latents $\mathbf{z}_t$. This allows the model to leverage the conditioning information implicitly without architectural modifications.

\subsection{Analysis:
Token and Computation Redundancy in In-Context Conditioning }\label{sec:ana}

Before detailing our FullDiT2, we first analyze the computational overhead introduced by reference conditions within the ICC paradigm. 

The core computational bottleneck in DiTs is the self-attention mechanism, whose cost scales quadratically with the input sequence length. In a standard DiT, processing noisy latents of length $N_x$ over $T$ timesteps and $L$ layers incurs an attention cost of approximately $O(T \cdot L \cdot N_x^2)$. However, ICC drastically rise this cost. By concatenating reference condition tokens (length $N_c$) with noisy latents, the total sequence length becomes $N_{\text{total}} = N_x + N_c$. If $N_c \approx N_x$ (e.g., a reference video in video editing), $N_{\text{total}} \approx 2N_x$, escalating the attention cost to $O(T \cdot L \cdot 4N_x^2)$---a four-fold increase. For tasks with multiple dense references, like Video Re-Camera using a reference video ($N_r$) and camera trajectory ($N_t$), where $N_r \approx N_x$ and $N_t \approx N_x$, the length $N_{\text{total}} \approx 3N_x$ can lead to a nine-fold increase, $O(T \cdot L \cdot 9N_x^2)$. This substantial burden motivates a deeper look into the computational costs. Our analysis identifies two primary sources of redundancy which we aim to mitigate, related to the context token sequence length ($N_c$) and the diffusion computation process ($L, T$):

\textbf{Token redundancy Analysis:} Token redundancy refers to the information redundancy within the conditional tokens. We analyze on the video object editing and re-camera tasks (analysis for more tasks in the appendix) using 100 samples, where the context tokens include the latent of reference source video as well as the identity image. Figure~\ref{fig:video_redudant} (a) shows the absolute difference across the averaged tokens of each consecutive frame in the source videos, which presents highly long-tail patterns and reveals the significant redundancy of the context tokens. Such redundancy lead to wasted computation in the full self-attention operations. As shown in Figure~\ref{fig:video_redudant} (b), even 50\% of reference tokens has accounted for more than 85\% of the attention scores from the noisy video latents. We carefully examine the attention patterns layer-by-layer in diffusion process, finding that in ICC, the network blocks dynamically focus on distinct subsets of these reference tokens (Figure~\ref{fig:video_redudant}, (c)) during the diffusion process. These observations reveal the context token length $N_c$ can be reduced.

\textbf{Computation redundancy Analysis:} Computation redundancy refer to the inefficiencies arising from the repetitive bi-directional full-attention computations between context tokens and noise latent across timesteps $T$ and layers $L$. Unlike the continuously updated noisy latents, our analysis reveals that the context tokens in ICC are relatively static across diffusion steps, as shown in (Figure~\ref{fig:video_redudant}, (d)). Continuously updating these stable context features leads to significant temporal computational redundancy, yet they could be effectively cached and reused. Moreover, such computation redundancy of clean context tokens not only happens across steps, but further across layers. To investigate the layer-wise redundancy, we measured the impact of context conditions on the noisy latent per layer with the block importance index (BI)\cite{sreenivas2024llm}. The results (Figure\ref{fig:video_redudant}, (e)) reveal a clear unbalanced pattern, indicating that the context tokens show significantly different levels of influence across layers, while the repetitive attention neglects this and bring less effective computations.

    


\subsection{
Efficient In-context Conditioning via Dynamic Token Selection and Selective Caching}
\begin{figure}[h]
    \centering
    \includegraphics[width=\linewidth]{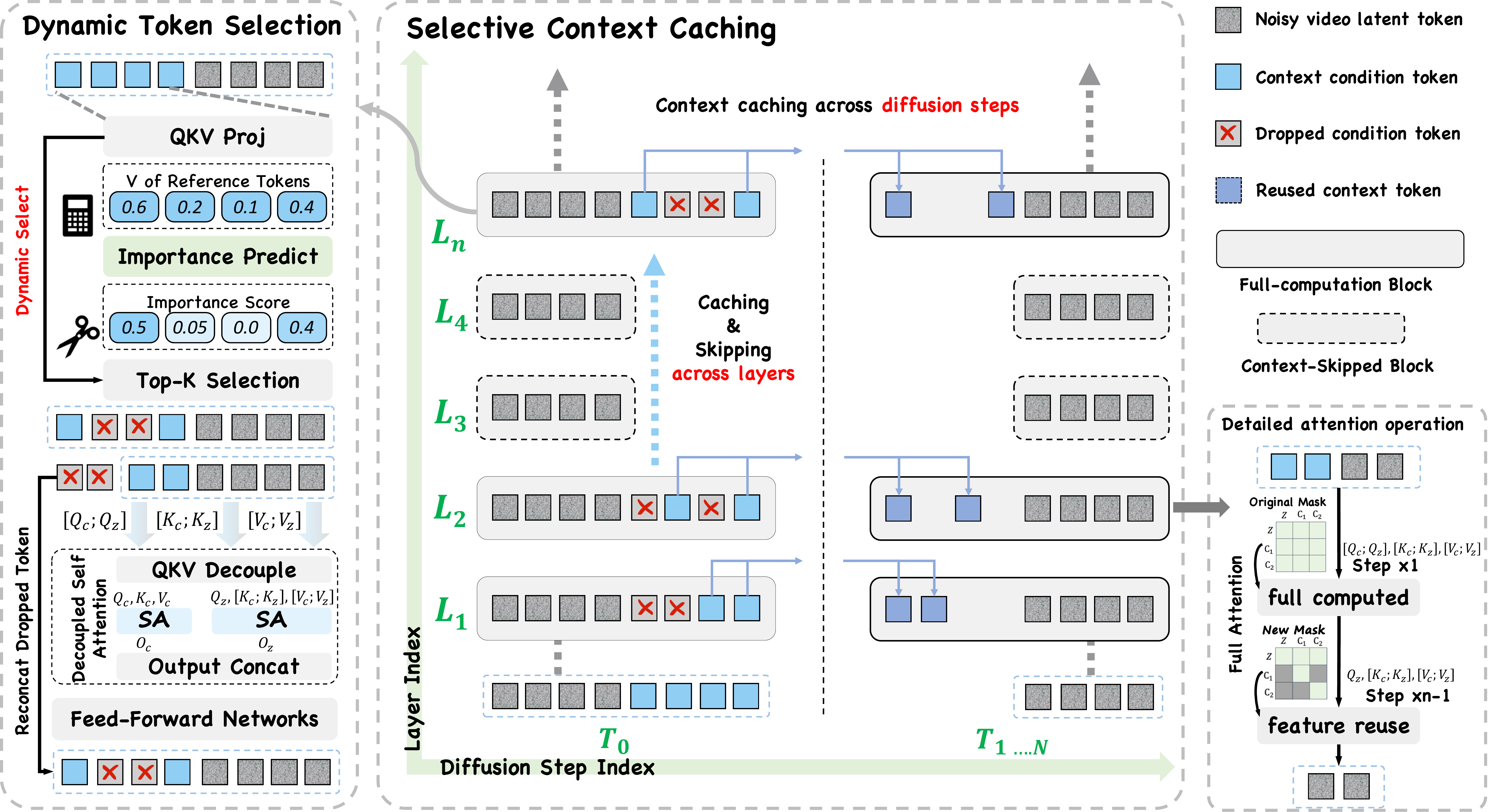}
    \caption{Overview of the FullDiT2 Framework. Left: The Dynamic Token Selection (DTS) module operates within each Transformer block at step $T_0$. It projects input hidden states to QKV, uses features derived from reference Value vectors ($V_c$) to predict importance scores via an MLP, and selects the top-K reference tokens. Selected tokens proceed to self-attention, while dropped tokens are re-concatenated after the FFN layer. Right: The Selective Context Caching mechanism illustrates temporal-layer caching. At $T_0$, DTS determines selected reference tokens. For subsequent steps like $T_{k}$, Transformer Block reuse cached K/V from selected reference tokens of $T_0$, while Skipped Blocks only process noisy tokens, significantly reducing computation.
}
    \label{fig:mainfig}
\vspace{-4mm}
\end{figure}

To mitigate the computational burden of dense reference conditions in In-Context Conditioning DiTs, we introduce FullDiT2, illustrated in Figure~\ref{fig:mainfig}. FullDiT2 optimizes conditioning token processing via two core mechanisms: Dynamic Token Selection, which adaptively selects a important subset of reference tokens per block to reduce reference token, and Selective Context Caching, which efficiently reuses selected reference K/V representations across timesteps and enables cross-layer reference token reuse, thereby minimizing redundant computations.

\subsubsection{Reducing Context Token Redundancy through Dynamic Selection}
Our analysis (Figure~\ref{fig:video_redudant}) reveals significant redundancy in reference tokens and varying layer-wise importance. To address this, each Transformer block is equipped with a Dynamic Token Selection module that adaptively selects an informative subset of reference tokens for attention processing.

Let the input hidden state to an attention block be $\mathbf{H}_{\text{in}} \in \mathbb{R}^{b \times n \times d}$. This is projected to Queries, Keys, and Values ($\mathbf{Q}, \mathbf{K}, \mathbf{V}$), which are then split into noisy latent components ($\mathbf{Q}_z, \mathbf{K}_z, \mathbf{V}_z \in \mathbb{R}^{b \times n_z \times d}$) and reference condition components ($\mathbf{Q}_c, \mathbf{K}_c, \mathbf{V}_c \in \mathbb{R}^{b \times n_c \times d}$).

Directly using attention scores from $\mathbf{Q}_z \mathbf{K}_c^T$ for selection is computationally expensive and often incompatible with optimized attention APIs~\cite{dao2022flashattention}. Inspired by prior work suggesting a correlation between Value vector norm and attention received~\cite{guo-etal-2024-attention}, we utilized a lightweight, learnable importance prediction network, $\phi$ (an MLP), to efficiently predict token salience. This network operates on features derived from the reference Value vectors, $f(\mathbf{V}_c)$, outputting a scalar importance score $\mathbf{s}_i$ for each reference token:
\begin{equation}
    \mathbf{s} = \phi(f(\mathbf{V}_c)) \in \mathbb{R}^{b \times n_c \times 1}
    \label{eq:dts_scores}
\end{equation}

Based on $\mathbf{s}$, we identify the indices $\mathcal{I}_{\text{topk}}$ of the top-$k$ scoring reference tokens (50\% tokens in our implementation). The attention mechanism then uses only this selected subset. We gather the corresponding reference QKV components: $\mathbf{Q}_{c, \text{topk}} = \text{gather}(\mathbf{Q}_c, \mathcal{I}_{\text{topk}})$, $\mathbf{K}_{c, \text{topk}} = \text{gather}(\mathbf{K}_c, \mathcal{I}_{\text{topk}})$, and $\mathbf{V}_{c, \text{topk}} = \text{gather}(\mathbf{V}_c, \mathcal{I}_{\text{topk}})$. Finally, we concate the selected reference token with noisy latent token:
\begin{equation}
\mathbf{Q}' = [\mathbf{Q}_z; \mathbf{Q}_{c, \text{topk}}], \quad
\mathbf{K}' = [\mathbf{K}_z; \mathbf{K}_{c, \text{topk}}], \quad
\mathbf{V}' = [\mathbf{V}_z; \mathbf{V}_{c, \text{topk}}]
\label{eq:dts_attention_inputs}
\end{equation}
This reduces the sequence length for attention involving reference tokens, lowering computational cost from $O((n_z + n_c)^2)$ towards $O((n_z + k)^2)$. To preserve information from unselected reference tokens (indexed by $\mathcal{I}_{\text{skip}}$) for subsequent layers, their original input hidden states $\mathbf{H}_{c, \text{skip}} = \text{gather}(\mathbf{H}_{\text{in}, c}, \mathcal{I}_{\text{skip}})$ (where $\mathbf{H}_{\text{in}, c}$ is the reference part of $\mathbf{H}_{\text{in}}$) bypass the attention mechanism. After the attention output is processed and passed through the block's Feed-Forward Network, yielding $\mathbf{H}_{\text{FFN}, [z;c,\text{topk}]}$, the bypassed $\mathbf{H}_{c, \text{skip}}$ are re-concatenated to form the complete output hidden state for the next block.

\subsubsection{Reducing Diffusion Computation Redundancy via Selective Context Caching}
Beyond token redundancy, the computational redundancy across timesteps $T$ and layers $L$ also brings huge computational costs. In response to this issue, we propose Selective Context Caching.

Given the reference token has different impact on noisy latent tokens across layers, we first identity the importance of layers on reference token processing by Block Importance Index~\cite{sreenivas2024llm}. For a set of $N_s$ samples, and for each candidate layer $l$, we compute two versions of the noisy latent output $\mathbf{O}_x^{(l)}$ from its attention module:
\begin{align}
    \mathbf{O}_{z, \text{no-ref}}^{(l)} &= \text{Attention}(\mathbf{Q}_z^{(l)}, \mathbf{K}_z^{(l)}, \mathbf{V}_z^{(l)}) \label{eq:skip_no_ref} \\
    \mathbf{O}_{z, \text{with-ref}}^{(l)} &= \text{Attention}(\mathbf{Q}_z^{(l)}, [\mathbf{K}_z^{(l)}; \mathbf{K}_c^{(l)}], [\mathbf{V}_z^{(l)}; \mathbf{V}_c^{(l)}]) \label{eq:skip_with_ref}
\end{align}
The Block Importance Index for layer $l$, $\text{BI}^{(l)}$, is then calculated as one minus the cosine similarity between these two outputs, averaged over the $N_s$ samples:
\begin{align}
    \text{BI}^{(l)} = \mathbb{E}_{\text{samples}} \left[ 1 - \frac{\mathbf{O}_{z, \text{no-ref}}^{(l)} \cdot \mathbf{O}_{z, \text{with-ref}}^{(l)}}{\|\mathbf{O}_{z, \text{no-ref}}^{(l)}\| \cdot \|\mathbf{O}_{z, \text{with-ref}}^{(l)}\|} \right] \label{eq:skip_bi}
\end{align}
A higher $\text{BI}^{(l)}$ indicates greater layer importance. There are 28 layers in our model,
we pre-select a fixed set of 4 layers with highest BI as "important layers" and first layer for token projection.
Only these important layers process reference information; the reference token representations output by one important layer are directly passed and reuse by subsequent important layer, bypassing any intermediate layers. These skipping layers only process noisy tokens.

Besides, repeatedly recomputing reference token at every timestep is inefficient due to the stable representation, as evidenced by OminiControl2~\cite{tan2025ominicontrol2}. 
However, naively caching and reusing $\mathbf{K}_{c,topk}, \mathbf{V}_{c,topk}$ across steps leads to training- inference misalignment, because reference queries ($\mathbf{Q}_{c,topk}$) would interact with noisy keys ($\mathbf{K}_z$) during training and bring unnecessary computation, as shown in the full attention in Figure~\ref{fig:mainfig}. To address this problem, an ideal solution is utilizing masked attention where $\mathbf{Q}_{c,topk}$ attends only to $\mathbf{K}_{c,topk}$, while noisy queries ($\mathbf{Q}_z$) attend to all keys ($[\mathbf{K}_z; \mathbf{K}_{c,topk}]$). As arbitrary masking can be inefficient or unsupported by optimized implementation like FlashAttention~\cite{dao2022flashattention}, we employ decoupled attention. This method splits the computation into two standard attention, replicating the desired mask attention result:
\begin{align}
    \mathbf{O}_{c,topk} &= \text{Attention}(\mathbf{Q}_{c,topk}, \mathbf{K}_{c,topk}, \mathbf{V}_{c,topk}) \label{eq:decoupled_Oc} \\
    \mathbf{O}_{z} &= \text{Attention}(\mathbf{Q}_{z}, [\mathbf{K}_z; \mathbf{K}_{c,topk}], [\mathbf{V}_z; \mathbf{V}_{c,topk}]) \label{eq:decoupled_Ox}
\end{align}
The attention output is $\mathbf{O} = [\mathbf{O}_x; \mathbf{O}_{c,topk}]$ (proof of equivalence in Appendix). 

During the inference, at first sampling step, for layers $l$ that are not skipped, we dynamically selected the important token and execute both attention computations (Eqs.~\ref{eq:decoupled_Oc} and \ref{eq:decoupled_Ox}) then cache the resulting reference $\mathbf{K}_{c,topk}^{(l)}$ and $\mathbf{V}_{c,topk}^{(l)}$ as $\mathbf{K}_{c,\text{cached}}^{(l)}, \mathbf{V}_{c,\text{cached}}^{(l)}$. 
At following steps, these same non-skipped layers, we only derive $\mathbf{Q}_z, \mathbf{K}_z, \mathbf{V}_z$ from the current noisy input $\mathbf{H}^{\text{in}}_{z}$. The Reference Self-Attention (Eq.~\ref{eq:decoupled_Oc}) is bypassed, and we execute only the Noisy-to-All Attention using the cached reference K/V to reduce the computational costs.

\begin{table}[ht]
\centering
\footnotesize 
\renewcommand{\arraystretch}{1.0} 
\caption{Quantitative Evaluation of FullDiT2 on ICC Tasks. Performance comparison of the baseline, baseline with individual FullDiT2 modules, and our full FullDiT2 framework across diverse tasks:
ID Insert/Swap/Delete, Video ReCamera, Trajectory-to-Video, and Pose-to-Video. 
Best results are highlighted in \textbf{bold}. $\uparrow$ indicates higher is better; $\downarrow$ indicates lower is better.}
\label{maintable_fitted}
\resizebox{\textwidth}{!}{
\begin{tabular}{@{}l ccc cc c ccc@{}} 
\toprule
\multicolumn{10}{c}{\textbf{ID Insert (V2V)}} \\
\midrule
\multicolumn{1}{c}{\multirow{2}{*}{\textbf{Method}}} & \multicolumn{3}{c}{\textbf{Efficiency}} & \multicolumn{2}{c}{\textbf{Identity}} & \multicolumn{1}{c}{\textbf{Alignment}} & \multicolumn{3}{c}{\textbf{Video Quality}} \\
\cmidrule(lr){2-4} \cmidrule(lr){5-6} \cmidrule(lr){7-7} \cmidrule(lr){8-10}
& \textbf{Lat.(step/s)$\downarrow$}& \textbf{GFLOPs$\downarrow$} & \textbf{SPEED$\uparrow$} & \textbf{CLIP-I$\uparrow$} & \textbf{DINO-S.$\uparrow$} & \textbf{CLIP-S.$\uparrow$}& \textbf{Smooth$\uparrow$}& \textbf{Dynamic$\uparrow$} & \textbf{Aesth.$\uparrow$} \\
\midrule
baseline & 0.533 & 69.292 & 1.000 & 0.568& 0.254& 0.227& 0.934& \textbf{17.576}& 5.372\\
baseline+Sel Caching: Step& 0.267 & 39.280 & 1.999 & 0.558& 0.249& \textbf{0.232}& \textbf{0.939}& 17.316& 5.328\\
baseline+Sel Caching: Layer& 0.300 & 38.851 & 1.776 & 0.597& 0.294& 0.231& 0.938& 16.497& 5.425\\
baseline+dyn. token sel. & 0.433 & 49.959 & 1.231 & 0.563& 0.257& 0.231& \textbf{0.939}& 16.850& \textbf{5.478}\\
ours & \textbf{0.233}& \textbf{33.141}& \textbf{2.287}& \textbf{0.605}& \textbf{0.313}& 0.229& 0.934& 17.286& 5.320\\
\midrule
\multicolumn{10}{c}{\textbf{ID Swap (V2V)}} \\
\midrule
baseline & 0.533 & 69.292 & 1.000 & 0.619& 0.359& 0.231& 0.932& 25.293& 5.312\\
baseline+Sel Caching: Step& 0.267 & 39.28 & 1.999 & 0.616& 0.353& 0.228& 0.934& \textbf{25.308}& 5.311\\
baseline+Sel Caching: Layer& 0.300 & 38.851 & 1.776 & \textbf{0.630}& \textbf{0.390}& 0.232& 0.932& 22.623& 5.323\\
baseline+dyn. token sel. & 0.433 & 49.959 & 1.231 & 0.625& 0.379& \textbf{0.233}& \textbf{0.935}& 24.647& \textbf{5.397}\\
ours & \textbf{0.233}& \textbf{33.141}& \textbf{2.287}& 0.621& 0.367& \textbf{0.233}& 0.929& 24.776& 5.233\\
\midrule
\multicolumn{10}{c}{\textbf{ID Delete (V2V)}} \\
\midrule
\multicolumn{1}{c}{\multirow{2}{*}{\textbf{Method}}} & \multicolumn{3}{c}{\textbf{Efficiency}} & \multicolumn{2}{c}{\textbf{Video Recon.}} & \multicolumn{1}{c}{\textbf{Alignment}} & \multicolumn{3}{c}{\textbf{Video Quality}} \\
\cmidrule(lr){2-4} \cmidrule(lr){5-6} \cmidrule(lr){7-7} \cmidrule(lr){8-10}
& \textbf{Lat.(step/s)$\downarrow$}& \textbf{GFLOPs$\downarrow$} & \textbf{SPEED$\uparrow$} & \textbf{PSNR$\uparrow$} & \textbf{SSIM$\uparrow$} & \textbf{CLIP-S.$\uparrow$}& \textbf{Smooth$\uparrow$}& \textbf{Dynamic$\uparrow$} & \textbf{Aesth.$\uparrow$} \\
\midrule
baseline & 0.533 & 69.292 & 1.000 & 27.432& 0.903& 0.214& 0.958& 9.702& 5.266\\
baseline+Sel Caching: Step& 0.267 & 39.28 & 1.999 & \textbf{27.748}& \textbf{0.912}& 0.214& 0.953& 10.211& 5.277\\
baseline+Sel Caching: Layer& 0.300 & 38.851 & 1.776 & 25.621& 0.867& 0.213& 0.951& \textbf{12.765}& \textbf{5.412}\\
baseline+dyn. token sel. & 0.433 & 49.959 & 1.231 & 26.014& 0.880& 0.218& \textbf{0.959}& 9.552& 5.216\\
ours & \textbf{0.233}& \textbf{33.141}& \textbf{2.287}& 26.236& 0.875& \textbf{0.221}& 0.953& 11.440& 5.333\\
\midrule
\multicolumn{10}{c}{\textbf{Video ReCamera (V2V)}} \\
\midrule
\multicolumn{1}{c}{\multirow{2}{*}{\textbf{Method}}} & \multicolumn{3}{c}{\textbf{Efficiency}} & \multicolumn{2}{c}{\textbf{Camera Ctrl.}} & \multicolumn{1}{c}{\textbf{Alignment}} & \multicolumn{3}{c}{\textbf{Video Quality}} \\
\cmidrule(lr){2-4} \cmidrule(lr){5-6} \cmidrule(lr){7-7} \cmidrule(lr){8-10}
& \textbf{Lat.(step/s)$\downarrow$}& \textbf{GFLOPs$\downarrow$} & \textbf{SPEED$\uparrow$} & \textbf{RotErr$\downarrow$} & \textbf{TransErr$\downarrow$} & \textbf{CLIP-S.$\uparrow$}& \textbf{Smooth$\uparrow$}& \textbf{Dynamic$\uparrow$} & \textbf{Aesth.$\uparrow$} \\
\midrule
baseline & 0.800& 101.517 & 1.000 & 1.855& 6.173& 0.222& 0.924& 26.952& 4.777\\
baseline+Sel Caching: Step& 0.333& 38.791 & 2.402& 1.590& \textbf{5.244}& 0.221& \textbf{0.931}& 27.625& 4.810\\
baseline+Sel Caching: Layer& 0.333& 43.944 & 2.402& \textbf{1.541}& 6.351& \textbf{0.227}& 0.928& 27.783& 4.812\\
baseline+dyn. token sel. & 0.433 & 64.463 & 1.924 & 1.802& 5.428& 0.223& 0.926& 29.266& 4.767\\
ours & \textbf{0.233}& \textbf{33.407} & \textbf{3.433}& 1.924& 5.730& 0.224& 0.923& \textbf{30.772}& \textbf{4.836}\\
\midrule
\multicolumn{10}{c}{\textbf{Trajectory to Video (T2V)}} \\
\midrule
baseline & 0.500& 64.457 & 1.000 & 1.471& 5.755& 0.214& 0.947& 23.209& 5.288\\
baseline+Sel Caching: Step& 0.267& 37.149 & 1.875& 1.596& 6.151& 0.219& 0.949& 22.589& \textbf{5.300}\\
baseline+Sel Caching:Layer& 0.267& 37.987 & 1.875& 1.438& 7.577& 0.210& \textbf{0.950}& 22.520& 5.249\\
baseline+dyn. token sel. & 0.433& 48.345 & 1.163& \textbf{1.391}& \textbf{5.528}& 0.217& 0.948& \textbf{23.779}& 5.267\\
ours & \textbf{0.233}& \textbf{33.111} & \textbf{2.143}& 1.566& 5.714& \textbf{0.221}& 0.943& 23.722& 5.211\\
\midrule
\multicolumn{10}{c}{\textbf{Pose to Video (T2V)}} \\
\midrule
\multicolumn{1}{c}{\multirow{2}{*}{\textbf{Method}}} & \multicolumn{3}{c}{\textbf{Efficiency}} & \multicolumn{2}{c}{\textbf{Pose Control}} & \multicolumn{1}{c}{\textbf{Alignment}} & \multicolumn{3}{c}{\textbf{Video Quality}} \\
\cmidrule(lr){2-4} \cmidrule(lr){5-6} \cmidrule(lr){7-7} \cmidrule(lr){8-10}
& \textbf{Lat.(step/s)$\downarrow$}& \textbf{GFLOPs$\downarrow$} & \textbf{SPEED$\uparrow$} & \multicolumn{2}{c}{\textbf{PCK$\uparrow$}} & \textbf{CLIP-S.$\uparrow$}& \textbf{Smooth$\uparrow$}& \textbf{Dynamic$\uparrow$} & \textbf{Aesth.$\uparrow$} \\
\midrule
baseline & 0.500& 64.457 & 1.000 & \multicolumn{2}{c}{\textbf{72.445}}& 0.244& 0.936& 17.238& 5.159\\
baseline+Sel Caching: Step& 0.267& 37.149 & 1.875& \multicolumn{2}{c}{71.982}& \textbf{0.247}& \textbf{0.939}& \textbf{18.442}& 5.163\\
baseline+Sel Caching: Layer& 0.267& 37.987 & 1.875& \multicolumn{2}{c}{71.775}& 0.246& 0.938& 17.156& \textbf{5.193}\\
baseline+dyn. token sel. & 0.433& 48.345 & 1.163& \multicolumn{2}{c}{71.702}& 0.242& \textbf{0.939}& 17.396& 5.178\\
ours & \textbf{0.233}& \textbf{33.111} & \textbf{2.143}& \multicolumn{2}{c}{71.408}& 0.246& \textbf{0.939}& 18.017& 5.174\\
\bottomrule
\end{tabular}}
\end{table}

\section{Experiment}
\subsection{Experiment Settings}
We evaluate FullDiT2 on diverse In-Context Conditioning (ICC) video tasks with dense references: Video Editing (ID Swap/Insert/Delete), Video Re-Camera~\cite{bai2025recammaster}, Pose-conditioned, and Camera trajectory-guided generation. Custom datasets were constructed for these tasks (details in Appendix). Efficiency is measured by latency, GFLOPs, and relative speedup. Quality is assessed using task-appropriate metrics including Aesthetic, Smoothness, and Dynamic scores~\cite{huang2024vbench}; CLIP Score (text-video alignment); DINO Score and CLIP-I (identity); PSNR/SSIM; RotErr/TransErr (camera control)~\cite{he2024cameractrl}; and PCK~\cite{gkioxari2014using}.
Due to limited prior work on optimizing reference conditions specifically for ICC video DiTs, our primary quantitative analysis compares baseline and different component of our methods applied on baseline.

\subsection{Comparison over Efficiency and Performance}
\textbf{Quantitative Comparisons:} Table~\ref{maintable_fitted} presents a quantitative comparison of our FullDiT2 framework against the baseline and individual FullDiT2 components across the diverse evaluation tasks. The results demonstrate a clear trend which incorporating individual components of FullDiT2 generally yields noticeable efficiency improvements with only a minor impact on generation quality metrics. Our full FullDiT2 framework, integrating all proposed optimizations, achieves the most significant efficiency gains while maintaining strong performance. For instance, in the ID-related video editing tasks, FullDiT2 achieves approximately 2.28x speedup over the baseline while preserving key quality metrics like identity similarity and overall video quality. The benefits are particularly pronounced in tasks with multiple conditions, such as Video ReCamera, which utilizes both a reference video and camera trajectory. In this scenario, FullDiT2 reduces the computational cost to only ~32\% of the baseline FLOPs and achieves a 3.43x speedup.
\begin{figure}[ht]
    \centering
    \includegraphics[width=\linewidth]{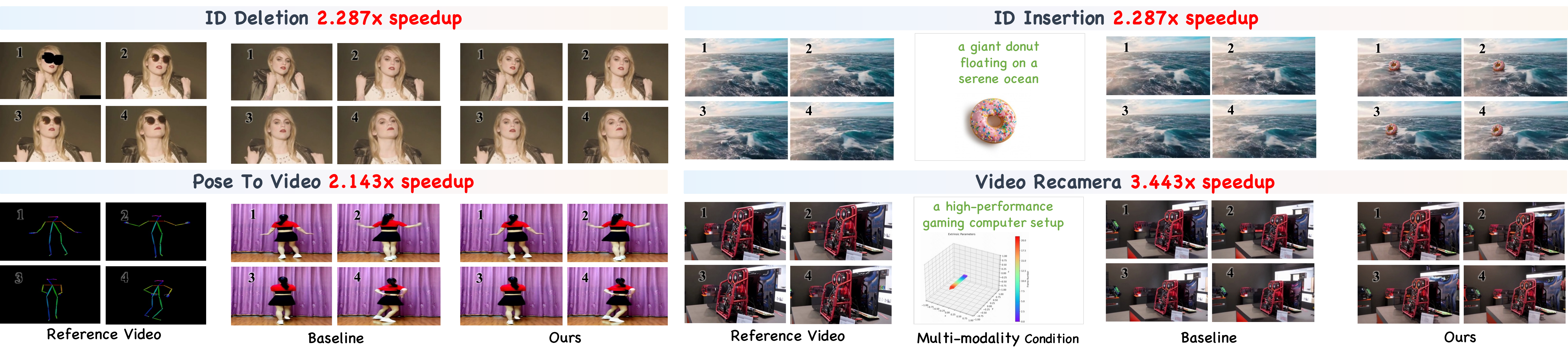}
    \caption{The qualitative comparison between FullDiT2 and baseline model on diverse tasks.}
    \label{fig:cmap}
\end{figure}

\textbf{Qualitative Comparisons:}
Figure~\ref{fig:cmap} presents qualitative comparisons of FullDiT2 against the baseline across diverse tasks; additional examples are in Appendix. The visual results demonstrate that FullDiT2 maintains high fidelity and accurately follow to various conditioning inputs and achieving results comparable to the baseline. In ID insertion tasks, FUllDiT2 even outperform baseline method. These examples qualitatively support our findings that FullDiT offers significant speedups while preserving essential visual quality and conditional consistency.



\subsection{Ablation Experiments}
\textbf{Analysis for context token selection ratio}
We validate the token redundancy mentioned in Figure~\ref{fig:video_redudant} by examining selection ratios and the importance of the selection strategy itself, with results in Table~\ref{tab:ablation_dts}. To determine an selection ratio, we experimented with ration of selected reference tokens. Results show that processing all tokens yields negligible quality benefits while bringing high computational cost, this confirms that a portion of reference tokens are redundant. However, aggressively dropping 80\% of tokens leads to significant performance degradation. Selecting approximately 50\% provides a efficiency-performance balance. To demonstrate that each reference tokens has different importance, we set the selection method into randomly selecting 50\% of tokens. 
The random approach resulted in poorer quality, demonstrating that a naive, uniform reduction in token count is insufficient.

\textbf{The effective of context caching across diffusion steps} 
We investigated the necessity and benefits of step-wise context caching strategy with results in Table~\ref{tab:ablation_dts}.
To assess the importance of maintaining training-inference consistency when caching, we replace decoupled attention with original full- attention during training. This configuration led to performance degradation in TransErr metric. This result demonstrate that Decoupled Attention is crucial for preventing training-inference misalignment. Next, to validate the efficiency gains and quality impact of step-wise context caching, we disabled this caching entirely, which increased latency by approximately 40\% and brought higher FLOPs, yet offered no quality benefits , even led to quality degradation, validating that recomputing reference token at every step is computationally expensive and unnecessary.


\begin{wrapfigure}{r}{8cm}
\vspace{-3mm}
    \includegraphics[width=\linewidth]{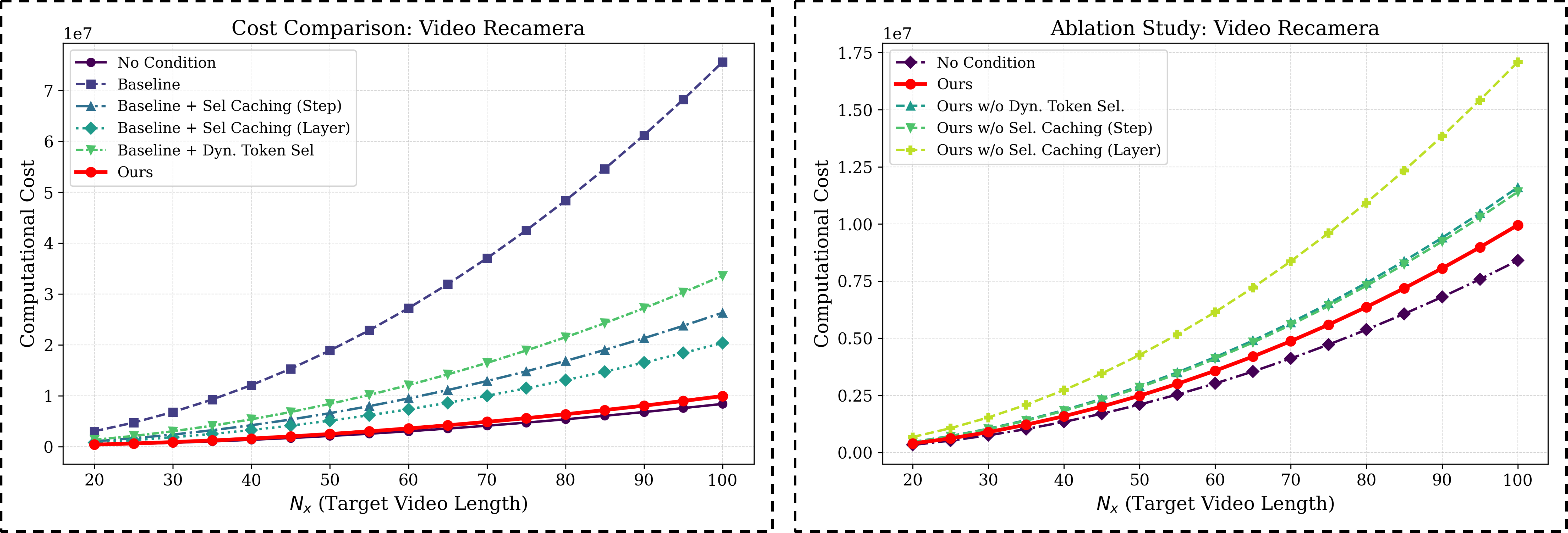}
    \caption{Comparison of computational costs of self-attention in different methods.}
    \label{fig:computation}
\vspace{-3mm}
\end{wrapfigure}

\textbf{Condition tokens show varying impact across layers} 
Our analysis (Figure~\ref{fig:video_redudant}) reveals the non-uniform, layer-dependent influence of reference tokens, quantified by the BI. To investigate this varying impact and validate our layer-wise context caching, we conducted three targeted experiments.
To begin, we removed layer-wise context caching, letting every layer to process reference tokens. This method yielded minimal quality benefits over context caching but brought higher computational costs. This demonstrates that computational redundancy exists across layers, and processing reference tokens in every layer is less efficient. However, relying solely on the single layer with the highest BI for reference processing was hard to maintain quality. To further evaluate the importance of selecting the specific layers, we experimented by retaining the same number of layers as our method but chose those with the lowest BI scores, resulting in markedly poorer performance. These results demonstrate the the condition tokens show varying impact across layers and the effectiveness of layer-wise selective context caching method.


\begin{table}[t] 
\centering
\small 
\renewcommand{\arraystretch}{1.1} 
\caption{Ablation Study of FullDiT2 Components on the Video ReCamera Task. 
We analyze the impact of each component and various design choices on efficiency and quality metrics.
}
\label{tab:ablation_dts} 
\setlength{\tabcolsep}{4pt} 
\begin{adjustbox}{width=\textwidth,center} 
\begin{tabular}{@{}llccccccccc@{}} 
\toprule
\textbf{Ablation} & \textbf{Variant} & \textbf{Latency(step/s)$\downarrow$}& \textbf{GFLOPS$\downarrow$}& \textbf{Speed$\uparrow$} & \textbf{CLIP-score$\uparrow$} & \textbf{Smoothness$\uparrow$} & \textbf{Dynamic$\uparrow$} & \textbf{Aesthetic$\uparrow$} & \textbf{RotErr$\downarrow$} & \textbf{TransErr$\downarrow$} \\
\midrule
Ours & - &  0.233&  33.407&  1.000&  0.224&  0.923&  30.772&  4.836&  1.924&  5.730\\
\midrule
\multirow{3}{*}{Dynamic Token Selection} & random selection &  0.233&  33.407&  1.000&  0.223&  0.932&  25.429&  4.812&  1.870&  6.923\\ 
 & drop rate 80\% &  0.233&  33.050&  1.000&  0.226&  0.927&  29.409&  5.040&  2.217&  8.887\\
 & Wo Dynamic Token Selection &  0.233&  33.992&  1.000&  0.225&  0.927&  26.968&  4.802&  2.238&  7.341\\
\midrule
\multirow{2}{*}{Sel Caching: Step}& Wo Decoupled Attention &  0.233&  33.407&  1.000&  0.218&  0.928&  26.588&  4.910&  1.607&  6.461\\ 
 & Wo Selective Context Caching: Step&  0.333&  37.987&  0.700&  0.226&  0.924&  29.829&  4.795&  1.962&  7.515\\ 
 \midrule
 \multirow{3}{*}{Sel Caching: Layer}& Preserve 1 Layer &  0.233&  32.467&  1.000&  0.223&  0.929&  22.848&  4.897&  3.908&  8.999\\
 & Preserve 4 lowest BI layers  &  0.233&  33.407&  1.000&  0.219&  0.878&  34.396&  4.857&  1.967&  7.638\\ 
 & Wo Selective Context Caching: Layer&  0.300
& 38.788&  0.777&  0.221&  0.932&  25.266&  4.834&  1.796&  6.877\\
\bottomrule
\end{tabular}
\end{adjustbox} 
\end{table}
\textbf{Analysis of Computational Cost:}
To further validate our method's efficiency, Figure~\ref{fig:computation} visualizes self-attention computational cost versus target video length, demonstrating FullDiT2 significant cost reduction compared to the baseline. We provide the detail analysis in appendix.


\section{Conclusion}
In this paper, we conducted a thorough analysis for the computation cost of the in-context conditioning framework, identifying two key sources of computational inefficiency. Based on the analytical results, we propose FullDiT2, an efficient in-context conditioning framework for controllable video generation and editing, featuring dynamic token selection to reduce token redundancy and a selective context caching mechanism to minimize computational redundancy of diffusion process. Extensive experiments demonstrate that FullDiT2 achieves significant acceleration over the baseline model while preserving comparable, high-quality video generation.

\section{Acknowledgment}
We thanks Ke Cao for the discussions in this work. 
\bibliographystyle{unsrt}
\bibliography{ref}





\newpage
\appendix
\section*{Appendix}

\section{Implementation Details}
\noindent\textbf{Base Model and Finetuning Setup.}
Our work builds upon the pretrained FullDiT model~\cite{ju2025fulldit}, which consists of 28 Transformer blocks. We initialize our FullDiT2 (and all compared variants) from this checkpoint and finetune all model parameters. Finetuning is conducted using the AdamW optimizer with a constant learning rate of $1 \times 10^{-5}$ for 400,000 iterations. Training was distributed across 32 NVIDIA 80GB GPUs. We trained ID-swap/insert/delete in a same checkpoint, and trained other tasks independently.

\noindent\textbf{Data Configuration.}
Input videos for training and inference are processed at a resolution of $672 \times 384$ pixels with a sequence length of 77 frames. A 3D VAE compresses these video sequences into latents of 20 temporal frames. The composition of input tokens varies by task:
\begin{itemize}
    \item \textbf{Video Editing (ID Swap, Insert, Delete):} Input consists of 20 frames of noisy video latents, 20 frames of reference video latents, and 3 frames representing the ID image. For ID Deletion, the ID image latents are replaced with zero tensors.
    \item \textbf{Pose-to-Video \& Trajectory-to-Video:} Input includes 20 frames of noisy video latents and 20 frames representing the pose sequence or camera trajectory respectively.
    \item \textbf{Video ReCamera:} Input comprises 20 frames of noisy video latents, 20 frames of reference video latents, and 20 frames representing the camera trajectory.
\end{itemize}

\noindent\textbf{FullDiT2 Configuration.}
\begin{itemize}
    \item \textbf{Dynamic Token Selection:} The module selects the top 50\% of reference tokens ($k=n_c/2$) per block for processing.
    \item \textbf{Selective Context Caching:}
        \begin{itemize}
            \item \textit{Step-wise Context Caching:} Reference Keys and Values (derived from selected tokens) are computed only at the first diffusion timestep ($t=0$) and are subsequently reused for all other timesteps.
            \item \textit{Layer-wise Context Caching:} We designate 5 layers for full reference processing. This includes the initial Transformer block (Layer 0), critical for early feature projection from input embeddings, and the four additional layers with the highest pre-computed Block Importance Indices (BI). Reference processing is skipped in the remaining 23 layers.
        \end{itemize}
\end{itemize}

\noindent\textbf{Inference and Evaluation.}
Inference speed metrics (latency, GFLOPs, speedup) are evaluated on the same GPU type used for training. All models  use 30 sampling steps for generation. Latency is reported as the average time per sampling step (total sampling time / 30). GFLOPs are measured by the overall GFLOPS/30.
For evaluation, ID-similarity metrics (DINO Score, CLIP-I) are computed between the provided reference ID image and each frame of the generated video. In video deletion tasks, PSNR and SSIM are calculated exclusively on the non-deleted (background) regions. The Percentage of Correct Keypoints (PCK) for pose accuracy is reported at a normalized threshold of 0.1.

\section{Dataset Details}
Our evaluation spans six distinct video conditioning tasks. For each task, we constructed dedicated training and testing datasets.
\subsection{ID Related}
\begin{figure}
    \centering
    \includegraphics[width=\linewidth]{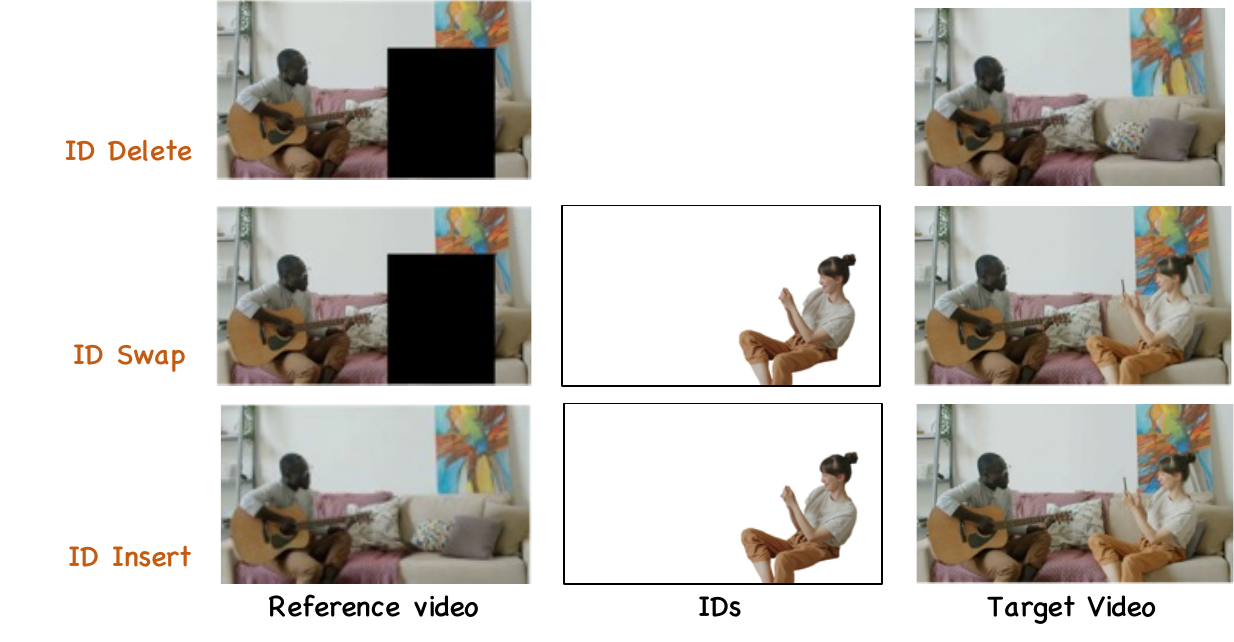}
    \caption{Training sample of our ID-related dataset.}
    \label{fig:supp-id-data}
\end{figure}

To generate training data for ID-related video editing tasks—namely, object deletion, swap, and insertion (demonstrated in Figure~\ref{fig:supp-id-data})—we employ a multi-stage pipeline.
First, for a given source video, we utilize SAM2~\cite{ravi2024sam} to automatically obtain a precise segmentation mask for the target object within each frame. This mask is then used with cv2.inpaint to create an inpainted version of the video where the object is naively removed. However, standard inpainting often introduces noticeable visual artifacts in the occluded regions.
To refine these inpainted results and produce artifact-free background videos, we train a dedicated ControlNet~\cite{zhang2023adding}. This ControlNet is conditioned on the original source video and guided by the inpainted video (or the mask itself) to learn to reconstruct the background seamlessly, effectively removing the inpainting artifacts.
The outputs of this pipeline serve as a basis for constructing our training triplets:
\begin{itemize}
\item For ID Insertion tasks, the artifact-free, inpainted video (object removed) serves as the \textit{reference video} (where a new object will be inserted). The original source video (containing the object to be "inserted" conceptually) acts as the \textit{ground truth video}.
\item For ID Deletion tasks, a masked version of the source video (where the target object is obscured by its segmentation mask) serves as the \textit{reference video}. The artifact-free, inpainted video (object successfully removed) acts as the \textit{ground truth video}.
\item For ID Swap tasks, the setup is similar to ID Deletion for the input . The \textit{ground truth video} would involve a different object inserted into the inpainted background, requiring an additional reference for the new object's identity.
\end{itemize}
Our evaluation set comprises 40 videos sampled from the total dataset for each task. Their captions were re-generated by GPT-4o and ensure that the specific video or ID were unseen during training.
\subsection{Pose guided video generation}
For the pose-conditioned video generation task, we constructed a dedicated dataset. The training set comprises approximately 160,000 video clips sourced from our internal human-centric video collection. For each video, we extracted a pose sequence using pose detection model. These extracted pose sequences serve as the conditioning signal during training and evaluation. The corresponding test set consists of 100 distinct videos with their derived pose sequences, held out from the training data. 

\subsection{Video Recamera and Trajectory to Video}
For these tasks, we employ the Multi-Cam Video dataset from the ReCamMaster~\cite{bai2025recammaster}. This established training dataset provides 136,000 videos.
For evaluation, we utilized $10$ basic camera trajectories and $50$ randomly selected videos from Koala~\cite{wang2024koala}. Each of the $10$ trajectories was then applied to $5$ distinct videos from this set (totaling $50$ trajectory-video pairs).

\section{Additional Experimental Results}
\subsection{Further analysis of Block Importance Index in Different tasks}

\begin{figure}
    \centering
    \includegraphics[width=\linewidth]{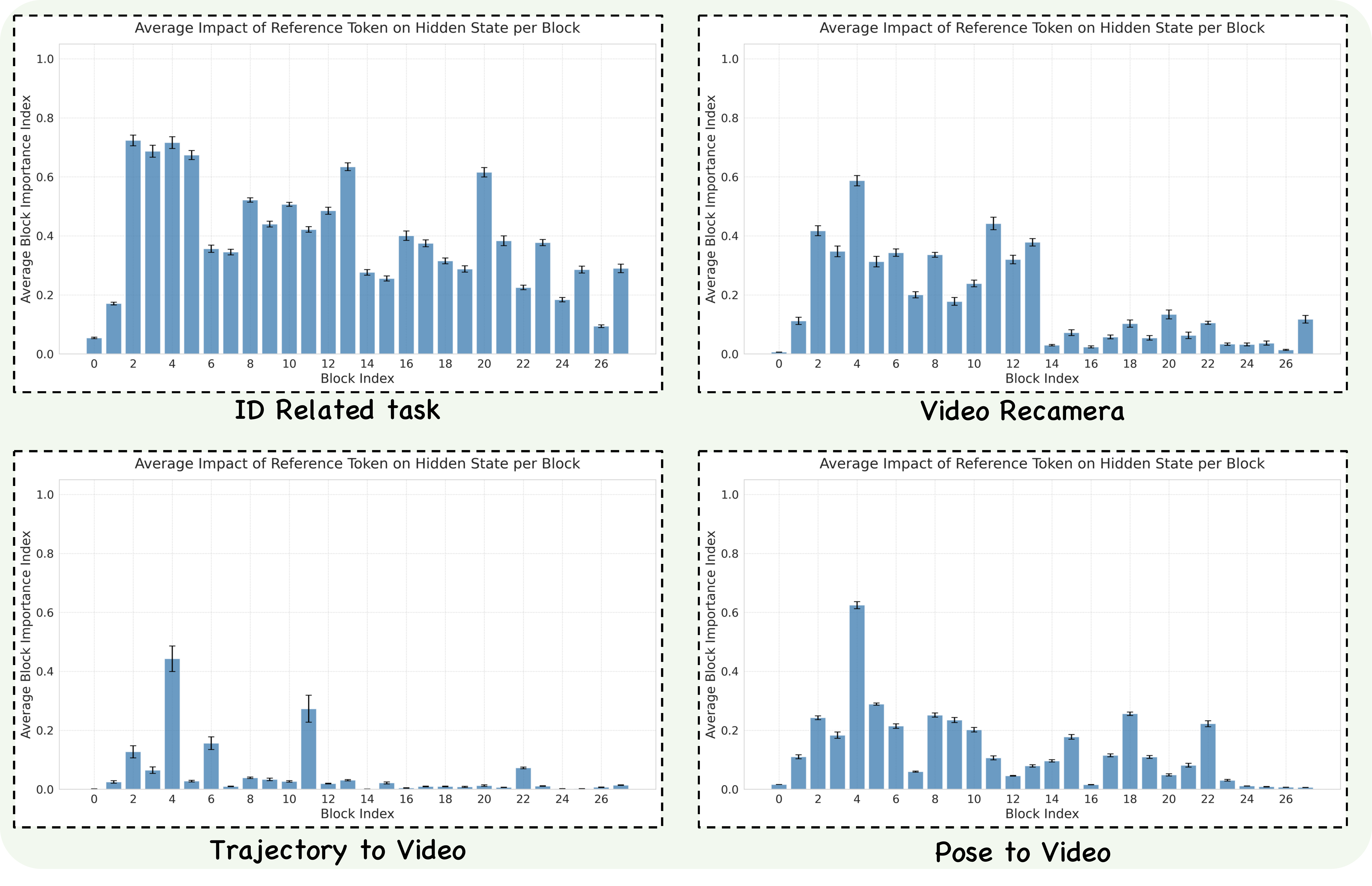}
    \caption{The block importance on different tasks.}
    \label{fig:blockana}
\end{figure}

In the main manuscript, we presented the Block Importance Index (BI) analysis for the Video ReCamera task. To further illustrate the generalizability of layer-wise variance in reference condition utilization, Figure~\ref{fig:blockana} extends this analysis to additional tasks, including ID-related task, pose-to-video and trajectory to video. The results reveal a non-uniform distribution of block importance across layers for these diverse tasks. This empirical evidence supports our strategy of pre-selecting a fixed set of high-importance layers for reference processing.

\begin{table}[ht]
\centering
\footnotesize 
\renewcommand{\arraystretch}{1.0} 
\caption{Quantitative Evaluation of FullDiT2 on ICC Tasks. Performance comparison of the baseline, baseline with individual FullDiT2 modules, and our full FullDiT2 framework across diverse tasks:
ID Insert/Swap/Delete, Video ReCamera, Trajectory-to-Video, and Pose-to-Video. 
Best results are highlighted in \textbf{bold}. $\uparrow$ indicates higher is better; $\downarrow$ indicates lower is better.}
\label{maintable_fitted}
\resizebox{\textwidth}{!}{
\begin{tabular}{@{}l ccc cc c ccc@{}} 
\toprule
\multicolumn{10}{c}{\textbf{ID Insert (V2V)}} \\
\midrule
\multicolumn{1}{c}{\multirow{2}{*}{\textbf{Method}}} & \multicolumn{3}{c}{\textbf{Efficiency}} & \multicolumn{2}{c}{\textbf{Identity}} & \multicolumn{1}{c}{\textbf{Alignment}} & \multicolumn{3}{c}{\textbf{Video Quality}} \\
\cmidrule(lr){2-4} \cmidrule(lr){5-6} \cmidrule(lr){7-7} \cmidrule(lr){8-10}
& \textbf{Lat.(step/s)$\downarrow$}& \textbf{GFLOPs$\downarrow$} & \textbf{SPEED$\uparrow$} & \textbf{CLIP-I$\uparrow$} & \textbf{DINO-S.$\uparrow$} & \textbf{CLIP-S.$\uparrow$}& \textbf{Smooth$\uparrow$}& \textbf{Dynamic$\uparrow$} & \textbf{Aesth.$\uparrow$} \\
\midrule
baseline & 0.533 & 69.292 & 1.000 & 0.568& 0.254& 0.227& 0.934& \textbf{17.576}& \textbf{5.372}\\
$\Delta-$DIT& 0.233& 34.688& 2.287& 0.519& 0.190& 0.217& \textbf{0.937}& 16.985& 5.253\\
FORA& \textbf{0.200}& \textbf{31.721}& \textbf{2.665}& 0.523& 0.184& 0.212& 0.932& 15.787& 5.352\\
ours & 0.233& 33.141& 2.287& \textbf{0.605}& \textbf{0.313}& \textbf{0.229}& 0.934& 17.286& 5.320\\
\midrule
\multicolumn{10}{c}{\textbf{ID Swap (V2V)}} \\
\midrule
baseline & 0.533 & 69.292 & 1.000 & 0.619& 0.359& 0.231& \textbf{0.932}& \textbf{25.293}& \textbf{5.312}\\
$\Delta-$DIT& 0.233& 34.688& 2.287& 0.561& 0.234& 0.224& 0.924& 26.047& 5.179\\
FORA& \textbf{0.200}& \textbf{31.721}& \textbf{2.665}& 0.588& 0.290& 0.224& 0.922& 25.053& 5.244\\
ours & 0.233& 33.141& 2.287& \textbf{0.621}& \textbf{0.367}& \textbf{0.233}& 0.929& 24.776& 5.233\\
\midrule
\multicolumn{10}{c}{\textbf{ID Delete (V2V)}} \\
\midrule
\multicolumn{1}{c}{\multirow{2}{*}{\textbf{Method}}} & \multicolumn{3}{c}{\textbf{Efficiency}} & \multicolumn{2}{c}{\textbf{Video Recon.}} & \multicolumn{1}{c}{\textbf{Alignment}} & \multicolumn{3}{c}{\textbf{Video Quality}} \\
\cmidrule(lr){2-4} \cmidrule(lr){5-6} \cmidrule(lr){7-7} \cmidrule(lr){8-10}
& \textbf{Lat.(step/s)$\downarrow$}& \textbf{GFLOPs$\downarrow$} & \textbf{SPEED$\uparrow$} & \textbf{PSNR$\uparrow$} & \textbf{SSIM$\uparrow$} & \textbf{CLIP-S.$\uparrow$}& \textbf{Smooth$\uparrow$}& \textbf{Dynamic$\uparrow$} & \textbf{Aesth.$\uparrow$} \\
\midrule
baseline & 0.533 & 69.292 & 1.000 & \textbf{27.432}& \textbf{0.903}& 0.214& \textbf{0.958}& 9.702& 5.266\\
$\Delta-$DIT& 0.233& 34.688& 2.287& 24.836& 0.858& 0.213& 0.944& 9.507& 5.281\\
FORA& \textbf{0.200}& \textbf{31.721}& \textbf{2.665}& 25.421& 0.872& 0.210& 0.940& 8.535& \textbf{5.384}\\
ours & 0.233& 33.141& 2.287& 26.236& 0.875& \textbf{0.221}& 0.953& \textbf{11.440}& 5.333\\
\midrule
\multicolumn{10}{c}{\textbf{Video ReCamera (V2V)}} \\
\midrule
\multicolumn{1}{c}{\multirow{2}{*}{\textbf{Method}}} & \multicolumn{3}{c}{\textbf{Efficiency}} & \multicolumn{2}{c}{\textbf{Camera Ctrl.}} & \multicolumn{1}{c}{\textbf{Alignment}} & \multicolumn{3}{c}{\textbf{Video Quality}} \\
\cmidrule(lr){2-4} \cmidrule(lr){5-6} \cmidrule(lr){7-7} \cmidrule(lr){8-10}
& \textbf{Lat.(step/s)$\downarrow$}& \textbf{GFLOPs$\downarrow$} & \textbf{SPEED$\uparrow$} & \textbf{RotErr$\downarrow$} & \textbf{TransErr$\downarrow$} & \textbf{CLIP-S.$\uparrow$}& \textbf{Smooth$\uparrow$}& \textbf{Dynamic$\uparrow$} & \textbf{Aesth.$\uparrow$} \\
\midrule
baseline & 0.800& 101.517 & 1.000 & 1.855& 6.173& 0.222& \textbf{0.924}& 26.952& 4.777\\
$\Delta-$DIT& 0.333& 50.758& 2.492& 1.798& 5.825& 0.219& 0.911& 26.066& 4.586\\
FORA& 0.300& 38.223& 2.667& \textbf{1.439}& 5.801& 0.218& 0.899& 24.998& 4.386\\
ours & \textbf{0.233}& \textbf{33.407}& \textbf{3.433}& 1.924& \textbf{5.730}& \textbf{0.224}& 0.923& \textbf{30.772}& \textbf{4.836}\\
\midrule
\multicolumn{10}{c}{\textbf{Trajectory to Video (T2V)}} \\
\midrule
baseline & 0.500& 64.457 & 1.000 & \textbf{1.471}& 5.755& 0.214& \textbf{0.947}& 23.209& \textbf{5.288}\\
$\Delta-$DIT& \textbf{0.200}& 32.240& \textbf{2.500}& 2.205& 7.093& 0.212& 0.942& 20.714& 5.030\\
FORA& \textbf{0.200}& \textbf{30.811}& \textbf{2.500}& 2.227& 7.848& 0.194& 0.938& 15.631& 4.512\\
ours & 0.233& 33.111& 2.143& 1.566& \textbf{5.714}& \textbf{0.221}& 0.943& \textbf{23.722}& 5.211\\
\midrule
\multicolumn{10}{c}{\textbf{Pose to Video (T2V)}} \\
\midrule
\multicolumn{1}{c}{\multirow{2}{*}{\textbf{Method}}} & \multicolumn{3}{c}{\textbf{Efficiency}} & \multicolumn{2}{c}{\textbf{Pose Control}} & \multicolumn{1}{c}{\textbf{Alignment}} & \multicolumn{3}{c}{\textbf{Video Quality}} \\
\cmidrule(lr){2-4} \cmidrule(lr){5-6} \cmidrule(lr){7-7} \cmidrule(lr){8-10}
& \textbf{Lat.(step/s)$\downarrow$}& \textbf{GFLOPs$\downarrow$} & \textbf{SPEED$\uparrow$} & \multicolumn{2}{c}{\textbf{PCK$\uparrow$}} & \textbf{CLIP-S.$\uparrow$}& \textbf{Smooth$\uparrow$}& \textbf{Dynamic$\uparrow$} & \textbf{Aesth.$\uparrow$} \\
\midrule
baseline & 0.500& 64.457 & 1.000 & \multicolumn{2}{c}{\textbf{72.445}}& 0.244& 0.936& 17.238& 5.159\\
$\Delta-$DIT& \textbf{0.200}& 32.240& \textbf{2.500}& \multicolumn{2}{c}{71.322}& 0.163& 0.919& 16.673& 4.946\\
FORA& \textbf{0.200}& \textbf{30.811}& \textbf{2.500}& \multicolumn{2}{c}{70.178}& 0.198& 0.936& \textbf{19.802}& 5.146\\
ours & 0.233& 33.111& 2.143& \multicolumn{2}{c}{71.408}& \textbf{0.246}& \textbf{0.939}& 18.017& \textbf{5.174}\\
\bottomrule
\end{tabular}}
\end{table}
\subsection{Comparison and discussion on general acceleration method}
We also compare FullDiT2 with general DiT acceleration techniques such as $\Delta$-DiT~\cite{chen2024delta}, FORA~\cite{selvaraju2024fora}. As shown in Table~\ref{maintable_fitted}, while these methods achieve noticeable latency reductions, they often incur a significant degradation in task- specific metric and overall quality when applied to our In-Context Conditioning tasks. This is likely because such methods primarily focus on reducing computational redundancy associated with the noisy latents, which, however, play a crucial role in maintaining high generation fidelity. In contrast, our approach targets redundancies specific to the reference condition signal.
\subsection{Analysis of Computational costs}
The primary computational bottleneck in Diffusion Transformers employing In-Context Conditioning is the self-attention mechanism. Its cost scales quadratically with the total input sequence length. We analyze the approximate computational cost, focusing on attention operations, for different configurations. Let $N_x$ be the sequence length of noisy latents, $N_c$ be the sequence length of reference condition tokens, $L$ be the total number of layers, $T$ be the number of diffusion steps, and $L_s$ be the number of layers actively processing reference tokens.

For clarity in illustrating the impact of dense conditions, we consider a scenario representative of our Video ReCamera task, where the reference condition length $N_c$ is approximately twice the noisy latent length, i.e., $N_c \approx 2N_x$. Consequently, after Dynamic Token Selection which halves the reference tokens, the effective reference length becomes $N'_c = \lfloor N_c/2 \rfloor \approx N_x$. The costs below are proportional to the number of token-pair interactions in attention.

\textbf{1. Baseline DiT (No Condition):}
In this standard configuration, no condition tokens are processed. Attention is computed only over the $N_x$ noisy latent tokens for all $L$ layers and $T$ timesteps.
\begin{equation}
 C_{\text{no\_cond}} = T L N_x^2
\end{equation}

\textbf{2. Baseline ICC (FullDiT-style):}
Here, the full set of $N_c$ reference condition tokens are concatenated with $N_x$ noisy latent tokens. All $L$ layers and $T$ timesteps perform attention over the total sequence length $(N_x + N_c) \approx 3N_x$.
\begin{equation}
 C_{\text{baseline\_icc}} \approx T L (3N_x)^2 = 9 T L N_x^2
\end{equation}

\textbf{3. FullDiT2 (Ours):}
Our method incorporates Dynamic Token Selection (reducing effective reference length to $N'_c \approx N_x$), step-wise context caching for reference tokens (computed at $t=1$, reused for $T-1$ steps), and layer-wise context caching ($L_s$ layers process reference, $L-L_s$ layers skip).

\textit{At the first timestep ($t=1$)}, reference K/V must be computed.
For the $L_s$ layers actively processing reference tokens, the approximate cost involves several components: $N_x(N_x+N_c)$ for noisy to all attention; $N'_c{}^2 \approx N_x^2$ for reference token self-attention. Summing these, each of these $L_s$ layers incurs a cost of roughly $3N_x^2$.
Concurrently, the $(L-L_s)$ layers that skip reference processing only perform noisy latent self-attention, costing $N_x^2$ each.
Thus, the total computational cost at $t=1$ is:
\begin{equation}
    C_{\text{ours}, t=1} \approx L_s (3N_x^2) + (L - L_s) N_x^2 = (L + 2L_s)N_x^2
    \label{eq:ours_t1_noitem}
\end{equation}

\textit{For each of the subsequent $(T-1)$ timesteps ($t>1$)}, cached reference K/V are reused.
In the $L_s$ layers actively processing reference tokens. The Noisy-to-All attention involves noisy queries $\mathbf{Q}_x$ attending to $[\mathbf{K}_x; \mathbf{K}'_{c, \text{cached}}]$; the component interacting with cached reference keys and noisy latent keys (length $N'_c \approx N_x$) costs approximately $N_x (N_x+N'_c) \approx 2N_x^2$.
The $(L-L_s)$ layers skipping reference processing maintain a cost of $N_x^2$ each.
Therefore, the total cost for each step $t>1$ is:
\begin{equation}
    C_{\text{ours}, t>1} \approx L_s (2N_x^2) + (L - L_s) N_x^2 = (L + L_s)N_x^2
    \label{eq:ours_t_gt_1_noitem}
\end{equation}
The total computational cost for FullDiT2 (Ours) aggregates these:
\begin{equation}
 C_{\text{ours}} = C_{\text{ours}, t=1} + (T-1)C_{\text{ours}, t>1} \approx (L + 2L_s)N_x^2 + (T-1)(L + L_s)N_x^2 = \left( TL + (T+1)L_s \right) N_x^2
 \label{eq:ours_total_noitem}
\end{equation}

\textbf{4. Baseline + Sel. Caching: Step:}
No Dynamic Token Selection is applied ($N_c \approx 2N_x$). All $L$ layers process reference tokens. Reference K/V are computed at $t=1$ and reused for the subsequent $T-1$ steps.
\begin{equation}
 C_{\text{step\_cache}} \approx L (7N_x^2) + (T-1) L (3N_x^2) = (3T + 4) L N_x^2
\end{equation}

\textbf{5. Baseline+Sel. Caching: Layer:}
No Dynamic Token Selection ($N_c \approx 2N_x$) and no step-wise context caching. Reference tokens are processed in $L_s$ layers, and skipped in $L-L_s$ layers, for all $T$ timesteps.
\begin{equation}
 C_{\text{layer\_cache}} \approx T \left( L_s (3N_x)^2 + (L - L_s) N_x^2 \right) = T(L + 8L_s)N_x^2
\end{equation}

\textbf{6. Baseline + Dynamic Token Selection:}
Only Dynamic Token Selection is active (effective reference length $N'_c \approx N_x$). All $L$ layers and $T$ timesteps process the reduced sequence length $(N_x + N'_c) \approx 2N_x$. No stepwise caching or layer caching.
\begin{equation}
 C_{\text{dts\_only}} \approx T L (N_x + N_x)^2 = 4 T L N_x^2
\end{equation}

\textbf{7. Ours w/o Dynamic Token Selection:}
FullDiT2 framework but with Dynamic Token Selection disabled ($N_c \approx 2N_x$ is used instead of $N'_c \approx N_x$). Context Caching are active.
\begin{equation}
 C_{\text{ours\_no\_dts}} \approx (L + 6L_s)N_x^2 + (T-1)(L + 2L_s)N_x^2 = \left( TL + (2T+4)L_s \right) N_x^2
\end{equation}

\textbf{8. Ours w/o Selective Caching: Step:}
FullDiT2 framework but with step-wise context caching disabled. Reference K/V are recomputed at every step. The cost of the first step is effectively repeated $T$ times.
\begin{equation}
 C_{\text{ours\_no\_temporal}} \approx T \left( L_s (3N_x^2) + (L - L_s) N_x^2 \right) = T(L + 2L_s)N_x^2
\end{equation}

\textbf{9. Ours w/o Selective Caching: Layer ($L_s \rightarrow L$):}
FullDiT2 framework but with layer-wise context caching disabled; all $L$ layers process reference tokens.
\begin{equation}
 C_{\text{ours\_no\_layer}} \approx 3 L N_x^2 + (T-1) 2 L N_x^2 = (2T + 1) L N_x^2
\end{equation}

\begin{table}[h!]
\centering
\caption{Approximate Computational Costs and Speedup Ratios vs. Baseline ICC (for $N_c \approx 2N_x$, $N'_c \approx N_x$). Costs are relative to $N_x^2$. Numerical speedups for $T=30, L=28, L_s=5$.}
\label{tab:comp_costs_speedup_numerical}
\renewcommand{\arraystretch}{2.2} 
\resizebox{\linewidth}{!}{
\begin{tabular}{@{}l >{\raggedright\arraybackslash}p{5.0cm} >{\centering\arraybackslash}p{4.0cm} >{\centering\arraybackslash}p{2.0cm}@{}}
\toprule
\textbf{Method} & \textbf{Approx. Cost Formula} & \textbf{Approx. Speedup Formula} & \textbf{Speedup} \\
                & \small ($C_{\text{method}} / N_x^2$) & \small ($9TL / (C_{\text{method}}/N_x^2)$) & \small ($T=30, L=28, L_s=5$) \\
\midrule
No Condition & $T L$ & $9 \times$ & $9.00 \times$ \\
\addlinespace
Baseline ICC (FullDiT) & $9 T L$ & $1.0 \times$ & $1.00 \times$ \\
\addlinespace
\textbf{FullDiT2 (Ours)} & $TL + (T+1)L_s$ & $\dfrac{9TL}{TL + (T+1)L_s}$ & $\approx 7.57 \times$ \\
\addlinespace
Baseline + Sel Caching: Step & $(3T + 4)L$ & $\dfrac{9T}{3T+4}$ & $\approx 2.87 \times$ \\
\addlinespace
Baseline + Sel Caching: Layer & $T(L + 8L_s)$ & $\dfrac{9L}{L+8L_s}$ & $\approx 3.71 \times$ \\
\addlinespace
Baseline + Dyn. Token Sel. & $4 T L$ & $2.25 \times$ & $2.25 \times$ \\
\addlinespace
Ours w/o Dyn. Token Sel. & $TL + (2T+4)L_s$ & $\dfrac{9TL}{TL + (2T+4)L_s}$ & $\approx 6.57 \times$ \\
\addlinespace
Ours w/o Sel Caching: Step & $T(L + 2L_s)$ & $\dfrac{9L}{L+2L_s}$ & $\approx 6.63 \times$ \\
\addlinespace
Ours w/o Sel Caching: Layer & $(2T + 1)L$ & $\dfrac{9T}{2T+1}$ & $\approx 4.43 \times$ \\
\bottomrule
\end{tabular}}
\end{table}

\textbf{Summary Table of Computational Costs and Speedup Ratios}
Finally, we summarize the approximate computational costs and their corresponding speedup ratios relative to the Baseline ICC method ($C_{\text{baseline\_icc}} \approx 9 T L N_x^2$). We assume the Video ReCamera scenario where $N_c \approx 2N_x$ and thus $N'_c = \lfloor N_c/2 \rfloor \approx N_x$. Numerical speedups are calculated using $T=30$, $L=28$, and $L_s=5$. As shown in Table~\ref{tab:comp_costs_speedup_numerical}, our method achieve a noticeable relative speedup of 7.57$\times$ compared with FullDiT.


\subsection{More visual results}
We provide more visual results in \textbf{index.html} in the zip.
\section{Proof of Equivalence for Decoupled Attention}
We propose a Decoupled Attention mechanism to enable efficient caching of reference Key ($\mathbf{K}_c$) and Value ($\mathbf{V}_c$) representations while preserving the intended information flow. This mechanism splits the attention computation into two parts. We demonstrate here that this Decoupled Attention is mathematically equivalent to a specific ideal masked attention pattern under standard scaled dot-product attention.

Let the query, key, and value projections for the noisy latent tokens be $\mathbf{Q}_z, \mathbf{K}_z, \mathbf{V}_z$, and for the reference/context tokens be $\mathbf{Q}_c, \mathbf{K}_c, \mathbf{V}_c$. The dimension of key vectors is $d_k$. We define the standard scaled dot-product attention as $\text{Attn}(Q, K, V) = \text{softmax}\left(\frac{QK^T}{\sqrt{d_k}}\right)V$.

Concatenate the queries, keys, and values into block matrices:
$$
\mathbf{Q}_{\text{full}} = \begin{bmatrix} \mathbf{Q}_z \\ \mathbf{Q}_c \end{bmatrix}, \quad
\mathbf{K}_{\text{full}} = \begin{bmatrix} \mathbf{K}_z \\ \mathbf{K}_c \end{bmatrix}, \quad
\mathbf{V}_{\text{full}} = \begin{bmatrix} \mathbf{V}_z \\ \mathbf{V}_c \end{bmatrix}
$$
Note that for the $\mathbf{K}_{\text{full}}^T$ operation, we would have $\mathbf{K}_{\text{full}}^T = [\mathbf{K}_z^T | \mathbf{K}_c^T]$.
The full unmasked attention score matrix $\mathbf{S}_{\text{full}}$ would be:
$$
\frac{\mathbf{Q}_{\text{full}}\mathbf{K}_{\text{full}}^T}{\sqrt{d_k}} =
\frac{1}{\sqrt{d_k}}
\begin{bmatrix} \mathbf{Q}_z \\ \mathbf{Q}_c \end{bmatrix}
[\mathbf{K}_z^T | \mathbf{K}_c^T] =
\frac{1}{\sqrt{d_k}}
\begin{bmatrix}
\mathbf{Q}_z \mathbf{K}_z^T & \mathbf{Q}_z \mathbf{K}_c^T \\
\mathbf{Q}_c \mathbf{K}_z^T & \mathbf{Q}_c \mathbf{K}_c^T
\end{bmatrix}
= \begin{bmatrix} \mathbf{S}_{zz} & \mathbf{S}_{zc} \\ \mathbf{S}_{cz} & \mathbf{S}_{cc} \end{bmatrix}
$$

\subsection{Ideal Masked Attention Pattern}
The ideal masked attention pattern requires that reference queries $\mathbf{Q}_c$ do not attend to noisy keys $\mathbf{K}_z$. This means the $\mathbf{S}_{cz}$ block in the score matrix is effectively masked to $-\infty$ before the softmax operation.
The masked score matrix $\mathbf{S}_{\text{masked}}$ becomes:
$$
\mathbf{S}_{\text{masked}} =
\begin{bmatrix}
\mathbf{S}_{zz} & \mathbf{S}_{zc} \\
-\infty & \mathbf{S}_{cc}
\end{bmatrix}
$$
Applying row-wise softmax:
Let $\mathbf{A}_{\text{masked}} = \text{softmax}(\mathbf{S}_{\text{masked}})$. Due to the $-\infty$ masking, the softmax will operate effectively independently on the rows corresponding to $\mathbf{Q}_z$ and $\mathbf{Q}_c$:
\begin{itemize}
    \item For rows corresponding to $\mathbf{Q}_z$:
    $$ \text{softmax}([\mathbf{S}_{zz} | \mathbf{S}_{zc}]) = [\mathbf{A}_{zz} | \mathbf{A}_{zc}] $$
    where $\mathbf{A}_{zz} + \mathbf{A}_{zc}$ (row sums) will be $\mathbf{1}$.
    \item For rows corresponding to $\mathbf{Q}_c$:
    $$ \text{softmax}([-\infty | \mathbf{S}_{cc}]) = [\mathbf{0} | \text{softmax}(\mathbf{S}_{cc})] = [\mathbf{0} | \mathbf{A}_{cc}] $$
    where $\mathbf{A}_{cc} = \text{softmax}(\mathbf{S}_{cc})$ (row sums will be $\mathbf{1}$).
\end{itemize}
So, the full attention weight matrix $\mathbf{A}_{\text{masked}}$ is:
$$
\mathbf{A}_{\text{masked}} =
\begin{bmatrix}
\mathbf{A}_{zz} & \mathbf{A}_{zc} \\
\mathbf{0} & \mathbf{A}_{cc}
\end{bmatrix}
$$
The output of the ideal masked attention $\mathbf{O}^{\text{masked}} = \mathbf{A}_{\text{masked}} \mathbf{V}_{\text{full}}$. However, since $\mathbf{V}_{\text{full}}$ is also block-structured, it's more common to consider the output components:
$$
\mathbf{O}^{\text{masked}} =
\begin{bmatrix}
\mathbf{O}_z^{\text{masked}} \\ \mathbf{O}_c^{\text{masked}}
\end{bmatrix}
=
\begin{bmatrix}
\mathbf{A}_{zz} & \mathbf{A}_{zc} \\
\mathbf{0} & \mathbf{A}_{cc}
\end{bmatrix}
\begin{bmatrix} \mathbf{V}_z \\ \mathbf{V}_c \end{bmatrix}
=
\begin{bmatrix}
\mathbf{A}_{zz}\mathbf{V}_z + \mathbf{A}_{zc}\mathbf{V}_c \\
\mathbf{A}_{cc}\mathbf{V}_c
\end{bmatrix}
$$
Thus,
\begin{align}
    \mathbf{O}_z^{\text{masked}} &= \mathbf{A}_{zz}\mathbf{V}_z + \mathbf{A}_{zc}\mathbf{V}_c \label{eq:block_ideal_oz} \\
    \mathbf{O}_c^{\text{masked}} &= \mathbf{A}_{cc}\mathbf{V}_c \label{eq:block_ideal_oc}
\end{align}

\subsection{Decoupled Attention Mechanism}
Our Decoupled Attention computes two outputs separately:
\begin{enumerate}
    \item \textbf{Reference Self-Attention:}
    \begin{equation}
        \mathbf{O}_c^{\text{decoupled}} = \text{Attn}(\mathbf{Q}_c, \mathbf{K}_c, \mathbf{V}_c) = \text{softmax}\left(\frac{\mathbf{Q}_c \mathbf{K}_c^T}{\sqrt{d_k}}\right) \mathbf{V}_c = \text{softmax}(\mathbf{S}_{cc}) \mathbf{V}_c = \mathbf{A}_{cc} \mathbf{V}_c
        \label{eq:block_decoupled_oc}
    \end{equation}

    \item \textbf{Noisy-to-All Attention:}
    Let $\mathbf{K}_{\text{all}} = [\mathbf{K}_z; \mathbf{K}_c]$ and $\mathbf{V}_{\text{all}} = [\mathbf{V}_z; \mathbf{V}_c]$.
    \begin{equation}
        \mathbf{O}_z^{\text{decoupled}} = \text{Attn}(\mathbf{Q}_z, \mathbf{K}_{\text{all}}, \mathbf{V}_{\text{all}})
        \label{eq:block_decoupled_oz_compact}
    \end{equation}
    The attention scores are $\text{softmax}\left(\frac{\mathbf{Q}_z \mathbf{K}_{\text{all}}^T}{\sqrt{d_k}}\right) = \text{softmax}([\mathbf{S}_{zz} | \mathbf{S}_{zc}]) = [\mathbf{A}_{zz} | \mathbf{A}_{zc}]$.
    Thus,
    \begin{equation}
        \mathbf{O}_z^{\text{decoupled}} = [\mathbf{A}_{zz} | \mathbf{A}_{zc}] \begin{bmatrix} \mathbf{V}_z \\ \mathbf{V}_c \end{bmatrix} = \mathbf{A}_{zz}\mathbf{V}_z + \mathbf{A}_{zc}\mathbf{V}_c
        \label{eq:block_decoupled_oz_expanded}
    \end{equation}
\end{enumerate}
The combined output of the Decoupled Attention (when both parts are computed) is implicitly $\mathbf{O}^{\text{decoupled}} = [\mathbf{O}_z^{\text{decoupled}}; \mathbf{O}_c^{\text{decoupled}}]$.

\subsection{Equivalence}
By comparing the derived components:
\begin{itemize}
    \item From Eq.~\eqref{eq:block_ideal_oc} and Eq.~\eqref{eq:block_decoupled_oc}: $\mathbf{O}_c^{\text{masked}} = \mathbf{A}_{cc}\mathbf{V}_c = \mathbf{O}_c^{\text{decoupled}}$.
    \item From Eq.~\eqref{eq:block_ideal_oz} and Eq.~\eqref{eq:block_decoupled_oz_expanded}: $\mathbf{O}_z^{\text{masked}} = \mathbf{A}_{zz}\mathbf{V}_z + \mathbf{A}_{zc}\mathbf{V}_c = \mathbf{O}_z^{\text{decoupled}}$.
\end{itemize}
Since both output components are identical, the Decoupled Attention mechanism correctly replicates the ideal masked attention pattern. This structure ensures that the computation of $\mathbf{O}_c$ (and thus the $\mathbf{K}_c, \mathbf{V}_c$ it depends on for caching) is isolated from the noisy latents, while $\mathbf{O}_z$ correctly gathers context from both noisy and reference tokens.

\section{Discussion on Limitations and Future Work}
While FullDiT2 demonstrates significant improvements in efficiency for In-Context Conditioning, we acknowledge certain limitations and areas for future exploration.

Firstly, our analyses identifying token and computational redundancies, as well as the specific configurations derived, are based on observations from the specific pretrained FullDiT model used in our experiments. Different base DiT architectures or models pretrained on vastly different data distributions might exhibit varying internal behavior. Consequently, the optimal hyperparameters for FullDiT2 components may not directly transfer and could require re-evaluation when applying our framework to other pretrained models. 

Secondly, the scope of our current task evaluation, while diverse, does not cover all potential ICC scenarios. The applicability and specific performance trade-offs of FullDiT2 in other domains, such as image-to-video generation, multi-shot video generation, or tasks with even longer and more complex concatenated conditions, need further investigation. 

\end{document}